\documentclass[conference]{ieeeconf}
\IEEEoverridecommandlockouts
\usepackage{cite}
\usepackage{amsmath,amssymb,amsfonts}
\usepackage{algorithmic}
\usepackage{graphicx}
\usepackage{textcomp}
\usepackage{tabularx}
\usepackage{multirow}
\usepackage{multicol}
\usepackage[table]{xcolor}
\usepackage{bbold}
\usepackage{hyperref}
\usepackage{float}
\hypersetup{
    colorlinks=true,
    linkcolor=blue,
    filecolor=magenta,      
    urlcolor=cyan,
    pdfpagemode=FullScreen,
}
\usepackage[font=small,belowskip=-1.5pt]{caption} 

\renewcommand{\baselinestretch}{0.97}
\renewcommand{\baselinestretch}{0.982}
\def\BibTeX{{\rm B\kern-.05em{\sc i\kern-.025em b}\kern-.08em
    T\kern-.1667em\lower.7ex\hbox{E}\kern-.125emX}}

\title{\LARGE \bf
SG-LSTM: Social Group LSTM for Robot Navigation\\ Through Dense Crowds
}

\author{Rashmi Bhaskara, Maurice Chiu, and Aniket Bera\\
{Department of Computer Science, Purdue University, USA}\\
Supplemental version including code and dataset at \small{\url{https://ideas.cs.purdue.edu/research/sg-lstm}}
\vspace{-10pt}
}

\begin{document}
\maketitle

\thispagestyle{empty}
\pagestyle{empty}

\begin{abstract}

As personal robots become increasingly accessible and affordable, their applications extend beyond large corporate warehouses and factories to operate in diverse, less controlled environments, where they interact with larger groups of people. In such contexts, ensuring not only safety and efficiency but also mitigating potential adverse psychological impacts on humans and adhering to unwritten social norms become paramount. In this research, we aim to address these challenges by developing a cutting-edge model capable of predicting pedestrian movements and interactions in crowded environments. To this end, we propose a novel approach called the Social Group Long Short-term Memory (SG-LSTM) model, which effectively captures the complexities of human group behavior and interactions within dense surroundings. By integrating social awareness into the LSTM architecture, our model achieves significantly enhanced trajectory predictions. The implementation of our SG-LSTM model empowers navigation algorithms to compute collision-free paths faster and with higher accuracy, particularly in complex and crowded scenarios. To foster further advancements in social navigation research, we contribute a substantial video dataset comprising labeled pedestrian groups, which we release to the broader research community. To thoroughly evaluate the performance of our approach, we conduct extensive experiments on multiple datasets, including ETH, Hotel, and MOT15. We compare various prediction approaches, such as LIN, LSTM, O-LSTM, and S-LSTM, and rigorously assess runtime performance.

\end{abstract}


\section{Introduction}

\begin{figure}[h]
    \centerline{\includegraphics[width=1\linewidth]{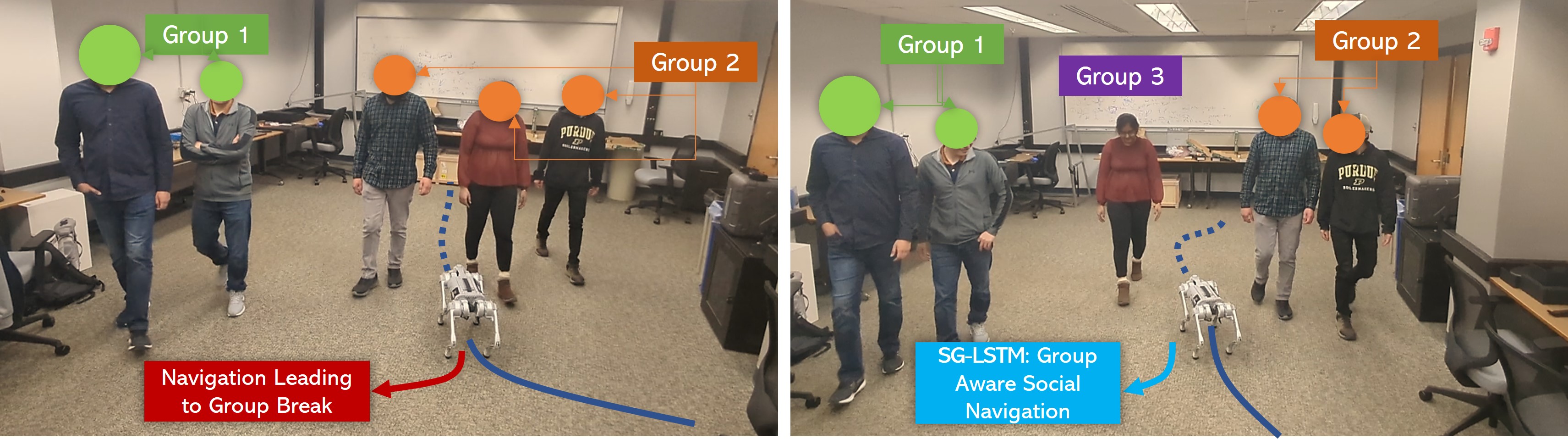}}
    \caption{\textit{Pedestrians in closer proximity are identified as groups. The robot will plan paths that avoid breaking pedestrian groups.}}
    \label{cover2}
\end{figure}

Social navigation is vital in social robotics as it enables robots to navigate and interact with human environments in a socially acceptable and effective way. Social navigation refers to the ability of a robot to move around in space while considering social norms and expectations. This includes navigating around people, avoiding obstacles, and following social conventions such as waiting in line or giving way to others. By incorporating social navigation into their design, social robots can move more smoothly and naturally through human environments, enhancing their social acceptance and effectiveness. This is particularly important for robots designed to interact with humans in public spaces, such as robots that provide guidance or assistance in airports, shopping centers, or hospitals. It also enables robots to interact with humans more effectively by allowing them to understand and respond to social cues and conventions. For example, a robot programmed to navigate a busy hospital corridor must recognize and avoid collisions with people walking in the opposite direction or understand when to yield the right of way to hospital staff rushing to an emergency. Existing socially compliant navigation algorithms are designed to assist service robots in navigating through complex environments safely while respecting social norms and avoiding collisions with pedestrians. However, some of these algorithms treat pedestrians as individual obstacles, making them unsuitable for use in crowded areas \cite{SC-MP-DRL, SC-Nav-Frozone, MA-CA-DRL-NC} where the social dynamics of pedestrians play a more significant role and impact on pedestrian dynamics. 
Fig.~\ref{cover2} demonstrates a problem that any navigation algorithm would face. A socially-aware navigation algorithm would choose to navigate around and not break a group, whereas a regular navigation algorithm that treats pedestrians as individual obstacles may cause the robot to freeze or decide to break a group. 

\begin{figure}[h]
    \centerline{\includegraphics[width=1\linewidth]{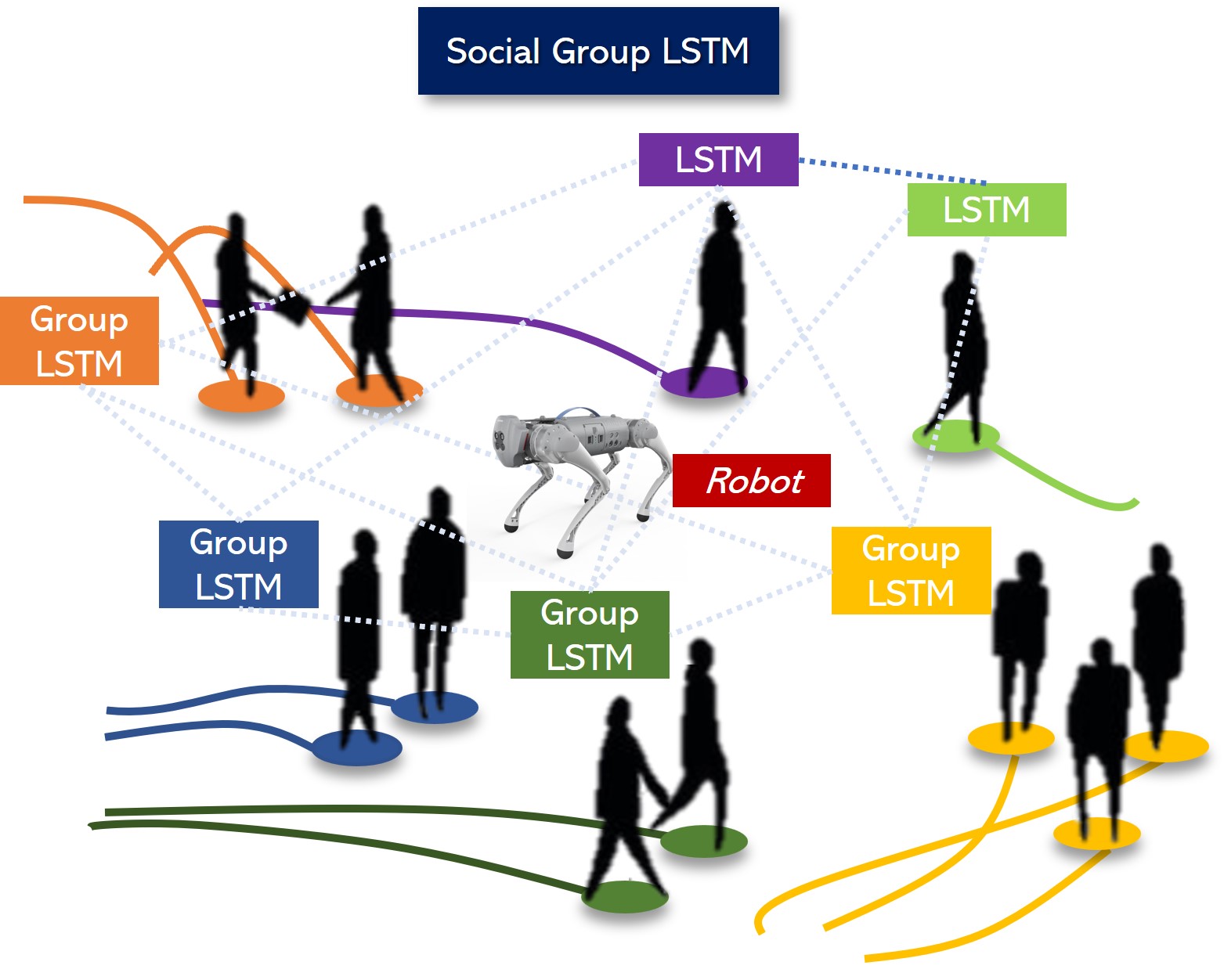}}
    \caption{\textit{\textbf{SG-LSTM: } We present a Socially-compliant, Group, Long Short-term Memory (SG-LSTM) model which leverages group dynamics into pedestrian prediction. Our hierarchical model decouples pedestrians on a group and individual level and can model the personal dynamics at both levels}.}
    \label{cover}
\end{figure}

In recent times, algorithms incorporating group cohesion have shown promising results in enhancing agents' navigation in dense-crowd environments. These algorithms rely on computing group cohesion, which involves analyzing factors like speed, direction, and proximity of individuals within a group to predict how closely they navigate together. However, these approaches have certain limitations, including the need for expensive online computations and reduced accuracy under specific conditions \cite{SC-Nav-CoMet}. High crowd density, for example, can significantly slow down the computation of multiple features, negatively affecting navigation performance. Additionally, the effectiveness of these algorithms is hindered when certain features cannot be detected, leading to poorly defined groups. To overcome these challenges, our paper presents an innovative approach that identifies perceptual groups in crowd videos employing a Long Short-term Memory (LSTM) architecture. This socially-compliant trajectory prediction algorithm improves navigation and is illustrated in Fig.~\ref{cover}. Our \textbf{main contributions} are as follows:
\begin{itemize}
    \item The Social Group Long Short-term Memory (SG-LSTM) model incorporates group dynamics into pedestrian prediction.
    \item Our hierarchical model independently models pedestrians at both group and individual levels, capturing personal dynamics effectively.
    \item Our group-optimized approach significantly reduces compute time by over  50\%.
    \item Additionally, we provide a large dataset containing more than 30,000+ human-labeled frames with detailed group information.
\end{itemize}

\noindent For the rest of the paper, we summarize some related work in Sec.~\ref{sec-relatedworks}. Sec.~\ref{sec-methodology} describes the overview and methodology of the system in detail. Followed by Sec.~\ref{sec-dataset}, that describes our dataset. In Sec.~\ref{sec-evaluation}, we discuss evaluation metrics, results, and analysis. Finally, Sec.~\ref{sec-conclusion} discusses future directions and limitations and concludes our approach. 

\section{Related Works}\label{sec-relatedworks}
This section reviews existing methods for group identification and pedestrian trajectory prediction.

\subsection{Human Group Learning}
Research has extensively focused on identifying group features and discovering groups using video input. Notably, \cite{Group-CrowdBehavior} adopts YOLOv2 for object detection, converting detections into feature vectors and clustering nearby individuals spatially. Objects are initially considered separate clusters, and groups are formed through iterative processes of computing Euclidean distances and updating centroids. Similarly, \cite{Group-IdentifyingSocialGroups} defines a social group based on proximity, speed, and movement direction of pedestrians. Additionally, Sathyamoorthy et al. \cite{SC-Nav-CoMet} propose mathematical models for pedestrian group characterization, considering walking speed, interaction, group size, and proximity as defining factors.

Apart from physical properties like positions and distances in crowds, group identification can also consider collective behaviors \cite{Group-GroupProfiling}. Shao et al. use the Collective Transition (CT) of pedestrians from tracklets to form clusters, defining groups based on high-velocity correlations with anchor tracklets. This approach provides insights into crowd dynamics and social groupings. These studies highlight diverse strategies, utilizing object detection, feature vectorization, clustering, and collective behavior analysis, to effectively identify and categorize groups in crowds from video data.

\subsection{Social Navigation}

This literature review explores a diverse range of research works that contribute to various aspects of efficient and safe navigation in complex scenarios. Cheung et al. \cite{cheung2012efficient} propose an innovative vehicle navigation approach based on driver behavior classification, while Chandra et al. \cite{chandra2016cmetric} introduce CMetric, a driving behavior measure utilizing centrality functions. Randhavane et al. \cite{randhavane2016pedestrian} present pedestrian dominance modeling for socially-aware robot navigation, considering pedestrians' dominance in crowded environments. Bera et al. \cite{bera2015adapt} develop Adapt, a real-time adaptive pedestrian tracking system, and Bera and Manocha \cite{bera2017realtime} present a real-time multilevel crowd-tracking method using reciprocal velocity obstacles. Additionally, Bera et al. \cite{bera2016glmp} introduce GLMP, a real-time pedestrian path prediction model incorporating both global and local movement patterns. Real-time anomaly detection in crowded scenes is addressed by Bera et al. \cite{bera2017realtimeanomaly}, and online parameter learning for data-driven crowd simulation and content generation is proposed by Bera et al. \cite{bera2017online}. The research works by Murino et al. \cite{murino2012group} explore group and crowd behavior analysis techniques within the domain of computer vision. Furthermore, Bera et al. \cite{bera2019socially} delve into the concept of ``The Socially Invisible Robot," aiming to enable robots to navigate and interact effectively in social environments. Additionally, Bera et al. \cite{bera2018interactive} present an interactive crowd-behavior learning system facilitating surveillance and training tasks. Chandra et al. \cite{chandra2017graphrqi} propose GraphRQI, a method for classifying driver behaviors using graph spectrums, and Cheung et al. \cite{cheung2012identifying} identify driver behaviors using trajectory features to enhance vehicle navigation and safety. These research works collectively contribute to advancing navigation and behavior analysis in various scenarios, providing valuable insights for the development of socially-aware robots and crowd management systems.

The field of computer vision has significantly contributed to modeling pedestrian behavior \cite{[1]}. The foundational Social Force Model proposed by Helbing and Molnar \cite{[1]} has found applications in crowd simulation \cite{[2]} and abnormal behavior detection \cite{[5]}. Further extensions include joint modeling of pedestrian trajectories and groupings \cite{jointModofPedAndGroups} and modeling social and group behavior for multiple people tracking \cite{neverWalkAlone}. Notably, there have been studies on people tracking with motion predictions from social forces \cite{[4]} and detecting social groups in video data \cite{[6]}. Additionally, image-based motion contexts have been explored for multiple people tracking \cite{[9]}. Recent works have explored diverse techniques for human activity prediction, including trajectory prediction with PORCA \cite{porca}, socially-aware navigation using SocioSense \cite{sociosense}, and modeling agent interactions with the LSTM-CNN hybrid network \cite{traphic}. Bera et al. \cite{glmp} predict pedestrian paths in complex environments. Studies also cover early activity recognition from video streams \cite{[42]}, recognizing human activities from partially observed videos \cite{[35]}, and predicting actions from static scenes \cite{[37]}. Data-driven activity prediction algorithms have been proposed \cite{[38]}, and Vondrick et al. \cite{[41]} present a deep learning-based framework for anticipating visual representations. Additionally, the research further explores pedestrian behaviors in stationary crowds, analyzing relationships \cite{[16]}, predicting saliency \cite{[17]}, and socially-aware forecasting \cite{[18]}. Techniques include data-driven analysis \cite{[20]}, Gaussian process regression flow \cite{[22]}, and trajectory learning \cite{[23]}. Predictive models cover behavior recognition \cite{[27]}, learning intentions \cite{[28]}, and action forecasting \cite{[29]}. Kitani et al. \cite{[30]} forecast future actions from visual input.

Current methods in social navigation often overlook the critical aspect of detecting and preserving the cohesion of social groups in a scene. By neglecting the presence and dynamics of such groups, these approaches may not achieve optimal prediction accuracy and fail to account for the social behavior of individuals. To address this limitation, we present our novel approach, SG-LSTM, which introduces a group-aware paradigm for trajectory prediction. Our approach demonstrates its effectiveness through extensive experiments on various datasets, showcasing its ability to predict socially-compliant trajectories while avoiding collisions with detected groups. By considering group dynamics in trajectory prediction, SG-LSTM represents a significant advancement in the field, promising more reliable and human-friendly robot navigation in complex real-world scenarios.

\section{Methodology} \label{sec-methodology}

\subsection{Overview}
\begin{figure}[ht]
    \centerline{\includegraphics[width=1\linewidth]{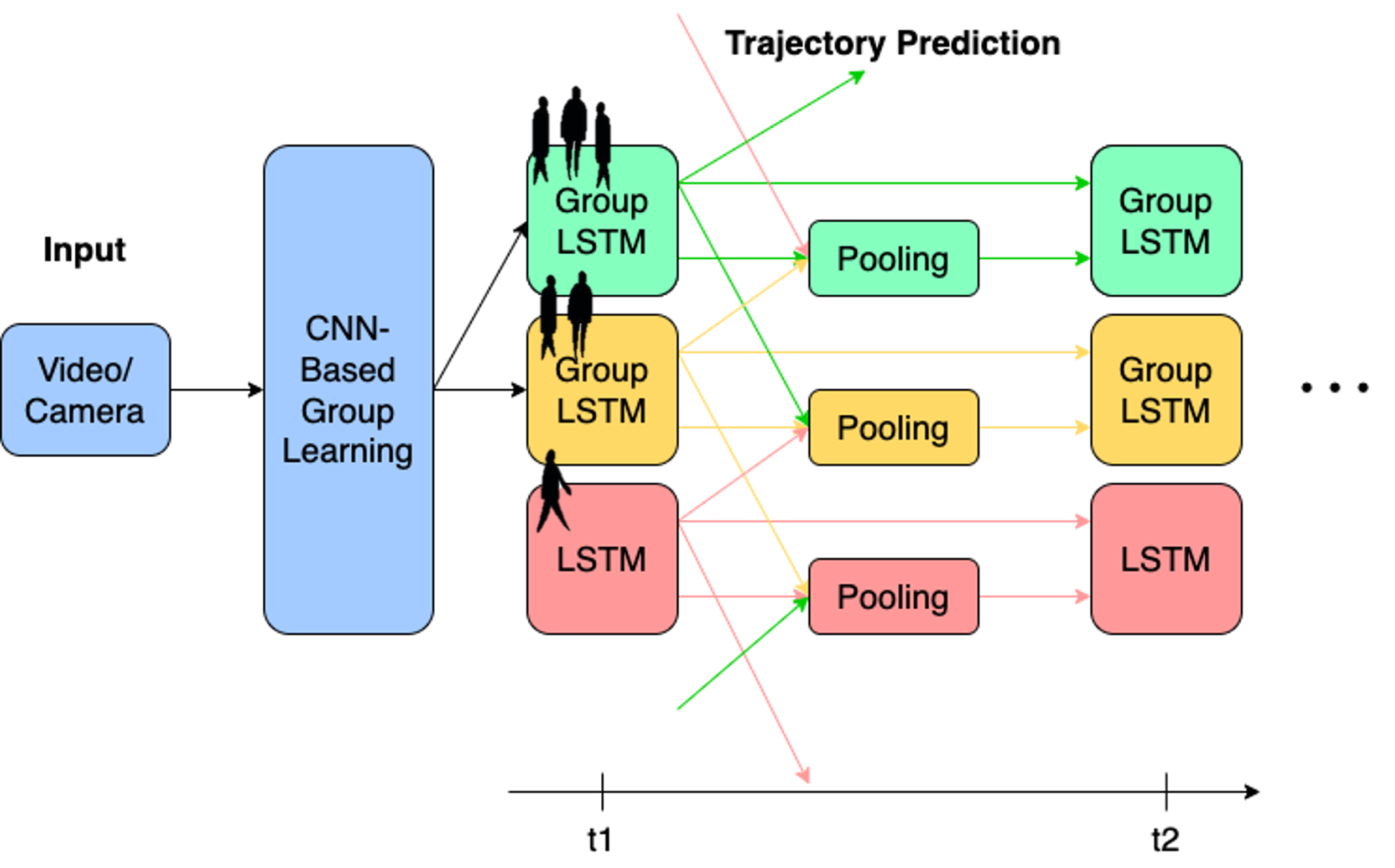}}
    \caption{Overview of our socially-compliant group, LSTM model. RGB frames captured from the camera will go through group detection so that spatial coordinates can be extrapolated and fed as input to the architecture.}
    \label{sg-lstm-arch}
\end{figure}

The SG-LSTM takes raw RGB and Depth frames from videos as input. Our CNN-based group learning algorithm then learns to predict groups in dense crowds, providing spatial coordinates of the groups for the trajectory prediction pipeline. The architecture is visualized in Fig.~\ref{sg-lstm-arch}, and all symbols used in this paper are summarized in Tab.~\ref{symbol_definitions}.
\vspace{-10pt}

\begin{table}[h]{}
\caption{Symbols and their definitions used in SG-LSTM}
\label{symbol_definitions}
\begin{center}
\begin{tabular}{|m{5em}|m{20em}|}
    \hline
    $d$ & Average depth of a group \\
    \hline
    $(x_b,y_b)$ & Centroid of the bounding box  \\
    \hline
    $FOV_{cam}$ & Field of view of the camera  \\
    \hline
    $w$ & Width of the image frame  \\
    \hline
    $\phi$ & Robot's orientation \\
    \hline
    $\phi_g$ & Angular displacement of a group w.r.t. the robot  \\
    \hline
    $h_{t}^{i}$ & The hidden layer of the LSTM of the $i^{th}$ group \\
    \hline
    $e_t^i$ & The vector embedding the input coordinates and hidden tensor of the $i^{th}$ group at time t \\
    \hline
    $(x_g, y_g)_t^i$ & The spatial coordinates of the $i^{th}$ group w.r.t. the robot at time t \\
    \hline
    $(x_r, y_r)_t$ & The position of the robot at time t \\
    \hline
    $u_s$ & Speed of the robot \\
    \hline
    $u_{\phi}$ & Steering angle of the robot \\
    \hline
\end{tabular}
\end{center}
\end{table}
\vspace{-15pt}

\vspace{-10pt}
\subsection{Trajectory Prediction Pipeline}

Our trajectory prediction algorithm builds upon the Social-LSTM model \cite{TP-SocialLSTM} designed for human trajectory prediction in crowded social scenes. Social-LSTM incorporates social interactions and past trajectories of individuals to predict future movement patterns. Although Social-LSTM is a powerful model for human trajectory prediction in crowded social scenes, it has limitations. Its accuracy tends to decrease with higher pedestrian density or when more agents form groups.  By introducing a superior group dynamics model, we enhance the state-of-the-art in two essential ways: achieving more accurate trajectory predictions and better resilience to inaccuracies in detection or sensor output. When we model the group, all pedestrians are considered within one group, allowing the prediction to rely on group trajectories even when tracking for some agents within the group is challenging. This robustness is especially valuable when pedestrian detection algorithms struggle to accurately extract spatial coordinates, particularly for people walking in groups or occluded scenes.

Scaling up Social-LSTM or similar models to handle large and complex scenes with numerous individuals presents challenges due to the substantial computational resources and memory required to capture the social context of the scene. As the model represents each person using an LSTM, a scene with multiple individuals necessitates a corresponding number of LSTMs, leading to increased computational demands. Moreover, these resources may be sub-optimally utilized when dealing with stationary people or individuals moving together as a group. Consequently, there is a growing need for more efficient and scalable trajectory prediction models that can handle such scenarios while maintaining prediction accuracy and computational efficiency.

\subsection{Problem Formulation}
In our approach, we establish spatial coordinates for each detected group as $(x_g, y_g)^i_t$ at any given time instant $t$. For pedestrians not belonging to any groups, we treat them as individual groups with a single member. These spatial coordinates are collected over five consecutive time steps, from $t_1$ to $t_5$, at a frame rate of 30 fps, and then fed into our Social-Group-LSTM model to predict trajectories for the subsequent five-time steps, spanning from $t_6$ to $t_{10}$. This time-step configuration allows us to effectively compute and forecast the movement patterns of both grouped and individual pedestrians, enhancing trajectory prediction accuracy and enabling socially-compliant navigation in crowded environments.

\subsection{Group Localization}
Our approach involves training a CNN-based group learning model on our annotated dataset. It builds on a single-stage object learning algorithm, utilizing a spatial pyramid pooling (SPP) module to detect objects at various scales and sizes. This capability proves beneficial for identifying groups with diverse shapes and sizes. Additionally, our approach effectively handles occlusions, where objects in a group partially or fully block each other through a combination of spatial features and attention mechanisms, ensuring accurate object detection within groups.

The trained model outputs bounding boxes around groups in crowded scenes. These group bounding boxes are combined with pedestrian bounding boxes for individuals not belonging to any group, which are already available from the dataset. The bounding boxes are then projected onto depth maps for each scene, and the depth of each pedestrian or group is computed using the centroid of the bounding box. By utilizing the pixel depth value ($d$), the frame width, and the camera's field of view angle ($FOV_{\text{cam}}$) in radians, we calculate spatial coordinates for all pedestrian groups and ungrouped pedestrians. The angular displacement of each pedestrian/group with respect to the robot can be determined using the centroid of the bounding box $(x_b, y_b)$ and the FOV of the camera, where $\phi_g = ({x_b}/{w}) \cdot FOV_{cam}$. The spatial coordinates of the pedestrians/groups are calculated as follows:$(x_g,y_g) = d\cdot[cos\phi_g, sin\phi_g]$.

\subsection{Modeling Social Interactions}

In our approach, we employ a social pooling layer, similar to Social-LSTM, to analyze group interactions in a scene. The pooling layer is defined by a tensor $H^i_t$ for the $i^{th}$ trajectory, which is given as:
\vspace{-8px}

\begin{equation}
H^i_t = \sum_{j \in \mathcal{N}i} 1{m,n}[(x_g,y_g)^j_t-(x_g,y_g)^i_t]h^j_{t-1}
\end{equation}
\vspace{-10px}

where $h^j_{t-1}$ represents the hidden state corresponding to the $j^{th}$ group at time step $t-1$, and $1_{m,n}$ is an indicator function that checks if the coordinates $(x_g,y_g)$ corresponding to the neighboring groups of the $i^{th}$ group, represented as a set $\mathcal{N}_i$, are present in the $(m,n)$ cell of the grid. This layer effectively captures the social interactions among all the groups in a scene.

The social pooling layer in the Social-Group LSTM model treats each group as a single entity when modeling pedestrian interactions in a scene. This advantageous approach makes it easier for the model to scale up to denser crowds as not every person in the background is represented by an LSTM, unlike the case in Social-LSTM. Consequently, this significantly reduces the number of parameters required to model a densely-crowded scene. Moreover, by grouping people together, the model overcomes any inaccuracies that arise from computing the spatial coordinates of occluded pedestrians in the background, further enhancing the overall accuracy of trajectory predictions.

\textbf{Model Layers}: Our model layers and weights are similar to the Social-LSTM model. The input coordinates and hidden tensor are embedded into the vector $e^i_t$ and are passed as input to the hidden state ($h^i_t$) of an LSTM cell with weight W, corresponding to the current time step of the $i^{th}$ group, $h^i_t = LSTM(h^i_{t-1},e^i_t;W)$.

\textbf{Trajectory Estimation:}
The predicted trajectories are assumed to belong to a normal distribution and thus are given as: $(\hat{x_g},\hat{y_g})^i_t \sim \mathcal{N}(\mu^i_t,\,\sigma^i_t, \,\rho^i_t)$. Furthermore, the model parameters are trained by minimizing the negative log-Likelihood loss for every $i^{th}$ trajectory.
\vspace{-6px}
\begin{equation}
L^i(W) = -\sum_{t_6}^{t_{10}}logP((x_g,y_g)^i_t|\mu^i_t,\,\sigma^i_t, \,\rho^i_t)
\end{equation}
\vspace{-6px}

The trajectories $(x_g,y_g)$ returned by the model are passed to a navigation system that can use this to compute optimal paths for a robot. Since we predict trajectories using social interactions between pedestrians employing a group detection model, we can say that the robots could leverage this to deploy a socially compliant navigation system that would avoid obstructing through groups.

\subsection{Robot Navigation}
\label{sec:navigation}

We deploy our algorithm on a Unitree robot dog. The algorithm for predicting paths outlined earlier can also be utilized to navigate through dense crowds or pedestrians without collisions. The approach is based on Generalized Velocity Obstacles (GVO)~\cite{WilkieGVO}. The combination of path prediction and GVO is used for car-like robots, taking into account their dynamic constraints.

In this context, we are employing kinematic constraints similar to a car and assuming that the robot can detect the location of dynamic obstacles, such as pedestrians and other robots, nearby, despite sensor noise. We have implemented this method on a robot to navigate a crowded walkway toward its intended destination.

The GVO navigation method is a technique that relies on velocity obstacles to navigate robots that have kinematic constraints. In our situation, we employ kinematic constraints similar to those of a car, and we assume that the robot can detect the positions of nearby moving obstacles, such as pedestrians, albeit with some noise. The robot uses our approach to anticipate the movement of each pedestrian as it navigates through the crowded walkway to reach its intended destination.
 
A conventional kinematic model \cite{WilkieGVO} is utilized for the robot. The robot's state is characterized by its position $(x_r,y_r)$ and orientation $\phi$. It controls the robot's speed and steering angle, represented by $u_s$ and $u_{\phi}$.


The configuration of the robot is expressed as its position $(x_r,y_r)$ and orientation $\phi$, and the robot has controls for speed and steering angle, represented by $u_s$ and $u_{\phi}$, respectively.

Assuming that the controls remain unchanged for a given time interval, the position of the robot at a specific time $t$ can be determined by the following expression:

The constraints for the robot are specified as follows:



\vspace{-6px}

\begin{equation}\label{eq:carpos}R(t, u) = \begin{pmatrix}\frac{1}{\tan(u_{\phi})}\sin(u_{s}\tan(u_{\phi})t) \\ -\frac{1}{\tan(u_{\phi})}\cos(u_{s}\tan(u_{\phi})t) +  \frac{1}{\tan(u_{\phi})} \end{pmatrix}.\end{equation}
\vspace{-6px}

For more details, please refer to \cite{WilkieGVO}.


The robot utilizes the prediction scheme we developed to anticipate the paths of the pedestrians and pedestrian groups and prevent any steering actions that could result in a collision. It is assumed that the pedestrians may not actively take measures to avoid colliding with the robot, which implies that the robot is responsible for ensuring collision avoidance.




\begin{figure}[b!]
    \centerline{\includegraphics[width=1\linewidth]{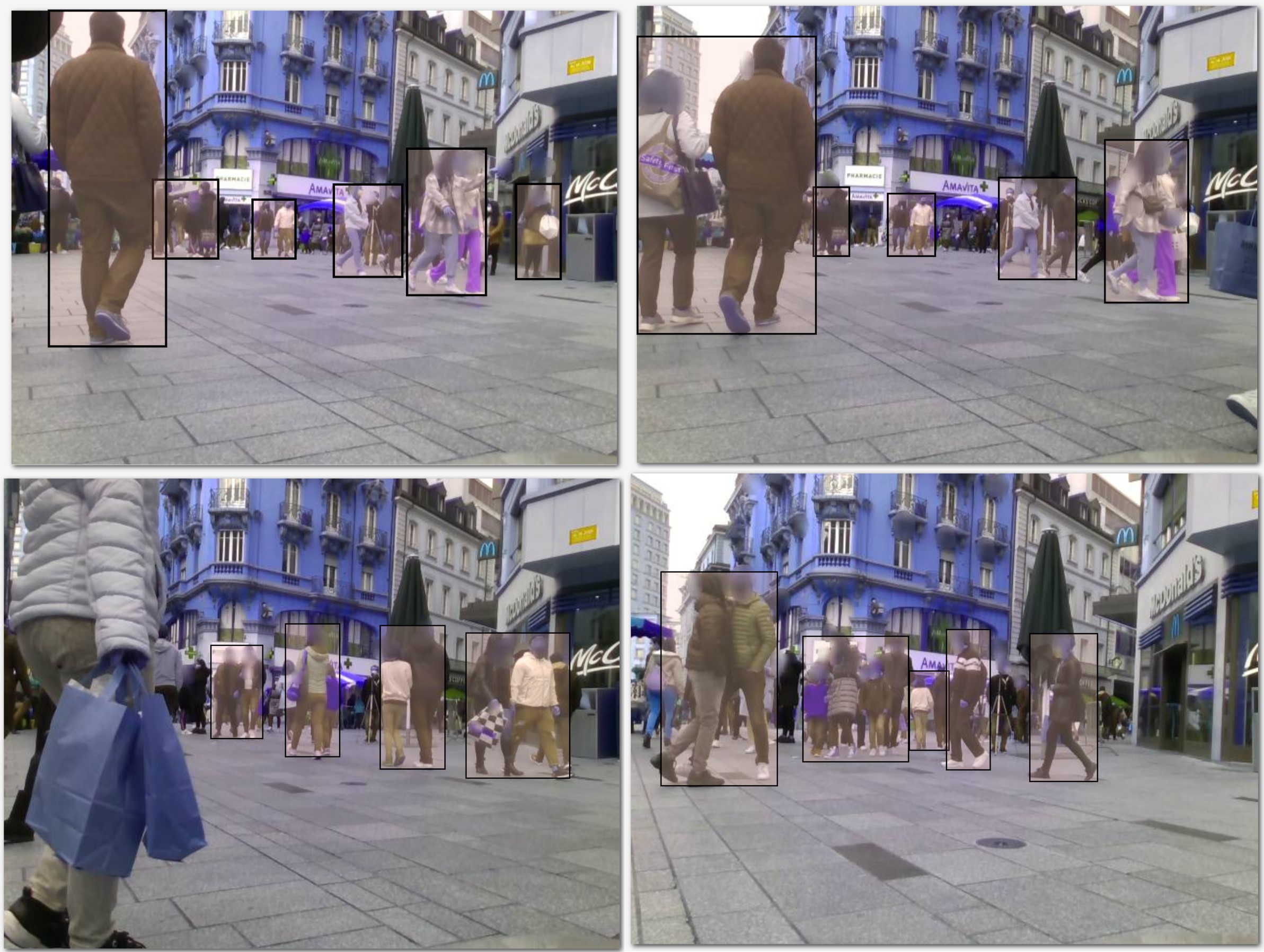}}
    \caption{Here, we briefly glimpse the pedestrian group labels (denoted by the bounding boxes) in our dataset.}
    \label{fig:g-ng}
\end{figure}

\section{Dataset}\label{sec-dataset}
\subsection{Our SG-LSTM Dataset and Data Annotation}
Our dataset (Fig. \ref{fig:g-ng}) is based on data collected on the Purdue Campus at West Lafayette and publicly available 3D point cloud, and RGB-D pedestrian dataset \cite{Dataset-PedRGBD}. In this study, we manually annotated perceptually-visible pedestrian groups in crowded scenes from RGB frames. We utilized a reference algorithm to aid in labeling, which involved processing RGB frames with 2D bounding boxes of detected pedestrians and their corresponding depth information. Pedestrians' positions were extrapolated relative to the robot's coordinates frame using methods presented in \cite{SC-Nav-CoMet}. Additionally, we calculated various features, such as group size, walking speed, and proximity of detected pedestrians, crucial for determining group cohesiveness. To classify pedestrians as a group, we referred to the conditions presented in \cite{SC-Nav-CoMet}. Subsequently, for all identified groups, we used CVAT.ai to accurately generate 2D bounding boxes. The labeled data was then used to train our model, following the methods explained in Sec.~\ref{sec-methodology}. The annotated dataset, which includes defaced, time-synced, color, and depth frames, along with the bounding boxes of pedestrian groups and individual pedestrians, will be released alongside this paper.

\begin{table}[h]
\caption{Quantitative results of all the methods on different datasets.}
\vspace{-10pt}

\label{evaluation}
\begin{center}
\begin{tabular}{|p{12em}|c|c|c|}
    \hline
    \multicolumn{1}{|c|}{Metric and Methods} & \multicolumn{3}{c|}{Datasets} \\
    \hline

    \textbf{Avg. Displacement Error} & \textbf{ETH} & \textbf{MOT15} & \textbf{Our Dataset}
    
    \\
    \hline
    Linear & 0.80 & 0.93 & 0.65 \\
    LSTM & 0.60 & 0.67 & 0.43 \\
    O-LSTM & 0.49 & 0.59 & 0.32 \\
    S-LSTM & 0.50 & 0.57 & 0.38 \\
    \textbf{SG-LSTM (Ours)} & \cellcolor{orange!25} 0.35 & \cellcolor{orange!25} 0.40 & \cellcolor{orange!25} \textbf{0.23} \\
    \hline
    \textbf{Final Displacement Error} & \textbf{ETH} & \textbf{MOT15} & \textbf{Our Dataset} \\
    \hline
    Linear & 1.31 & 1.01 & 0.91 \\
    LSTM & 1.31 & 0.70 & 0.58 \\
    O-LSTM & 1.06 & 0.66 & 0.41 \\
    S-LSTM & 1.07 & 0.69 & 0.42 \\
    \textbf{SG-LSTM (Ours)} & \cellcolor{orange!25} 0.68 & \cellcolor{orange!25} 0.48 & \cellcolor{orange!25} \textbf{0.27} \\
    \hline
\end{tabular}
\end{center}
\vspace{-20pt}
\end{table}


\section{Evaluation}\label{sec-evaluation}
In this section, we summarize the performance between our method and the other methods on the metrics chosen for evaluation on our SG-LSTM dataset, the ETH and Hotel dataset, and the MOT15 dataset. Below are the methods we have tested:
\begin{itemize}
  \item \textbf{Linear Model}: We assumed the pedestrian followed a linear path
  \item \textbf{LSTM}: Vanilla-LSTM with no social pooling and without grouping
  \item \textbf{O-LSTM}: LSTM with occupancy maps
  \item \textbf{S-LSTM}: LSTM with social pooling
  \item \textbf{SG-LSTM} (Ours): LSTM with social pooling and grouping
\end{itemize}

\noindent And, we used the following metrics to measure the accuracy:
\begin{itemize}
    \item \textbf{Average Displacement Error}: \\
    We calculate the mean squared error for all the predicted points in trajectories for every individual present in the scene as in \cite{neverWalkAlone}.
    \item \textbf{Final Displacement Error}: \\
    We compute the distance between the expected trajectory endpoint and the actual endpoint for every person.
\end{itemize}


\subsection{Displacement Errors of Predicted Trajectories}
By assuming that the pedestrians traveling as a group follow the same trajectory, we do not need to compute the trajectory for every individual pedestrian. Other trajectory-predicting methods that treat pedestrians as individuals would be less efficient in densely crowded environments due to needing to add trajectories for every person in the scene. And when these methods compute the trajectory for one person, it might treat the surrounding people as individual obstacles. In contrast, our model can avoid these expensive computations by knowing that these people belong to the same group. We show that our assumption that a group of pedestrians shares the same trajectory holds by calculating the average displacement error and the final displacement error between the predicted trajectories and the ground truth, where the ground truth is the trajectories of each person in the scene. The computed error in Tab.~\ref{evaluation} confirms that our assumption holds and does not affect the model's accuracy.

\subsection{Runtime Performance}
We observe our approach, \textbf{\textit{SG-LSTM}}, takes, on average, 55.4\% less time (Tab.~\ref{runtime}) than S-LSTM \cite{TP-SocialLSTM}. On average, 30-50\% of our tracked pedestrians belong to a group that significantly improves our compute time and is hence more appropriate for edge devices like robots. Lower compute requirements translate to lower costs for both hardware and software development. This means that robot navigation systems can be made more affordable and accessible to a wider range of users. The result is better performance, in terms of accuracy and speed, compared to Social-LSTM.

    
\begin{table}[t]
\centering
\caption{Average runtime for each  model to compute a trajectory  measured on a densely-crowded scene}
\label{runtime}
\centering
\begin{tabular}{|m{10em}|m{12em}|}
    
    \hline
    \textbf{Methods} & \textbf{Average Runtime (ms)} \\
    \hline
    Linear & 8.4 \\ 
    \hline
    S-LSTM \cite{TP-SocialLSTM} & 101 \\
    \hline
    O-LSTM \cite{TP-SocialLSTM} & 63 \\ 
    \hline
    Vanilla-LSTM & 57 \\ 
    \hline
    SG-LSTM (\textbf{Ours}) & \cellcolor{orange!25} 45 \\
    \hline
\end{tabular}
\vspace{-20pt}
\end{table}



\section{Conclusions, Limitations, and Future Work}\label{sec-conclusion}
We proposed a Social Group Long Short-term Memory (SG-LSTM) model which leverages group dynamics into pedestrian prediction. Our hierarchical model decouples pedestrians on a group and individual level and can model the personal dynamics at both levels. We also show that predicted trajectories can be used for a more socially-efficient navigation system that can leverage this to plan an optimal path for the robot. We also offer runtime numbers and demonstrate that our group-optimized approach leads to over 50\% reduction in compute time. 

At the same time, there are some limitations to our approach. Pedestrian prediction models like ours often rely solely on the position and motion of pedestrians without considering other contextual factors such as weather, time of day, or the presence of obstacles. Additionally, our model is designed to work only at short distances, such as within a few meters of the robot. This can limit the ability of the robot to plan for long-term interactions with pedestrians. Many pedestrian prediction models are complex and computationally intensive, making them difficult to deploy on resource-constrained robotic platforms.

In the future, we would like to work on these issues. Our model relies on data from a single sensor modality, such as cameras. Future work could explore integrating data from multiple modalities, such as vision, lidar, and radar, to improve the accuracy and robustness of pedestrian prediction models. In the future, we could consider a broader range of contextual information, such as weather conditions, time of day, and pedestrian behavior patterns. This could lead to more accurate and reliable predictions in various scenarios. We could explore how to extend these models to longer time horizons, such as predicting the trajectory of a pedestrian over the next minute or more.


\vspace{12pt}
\end{document}